# Graph Distance Neural Networks for Predicting Multiple Drug Interactions

**Haifan Zhou**

Ainnocence

zhouhf@connect.hku.hk

**Wenjing Zhou**

Ainnocence

zhouwenjing@mail.dlut.edu.cn

**Junfeng Wu\***

Ainnocence

junfeng.wu@ainnocence.com

## Abstract

Since multidrug combination is widely applied, the accurate prediction of drug-drug interaction (DDI) is becoming more and more critical. In our method, we use graph to represent drug-drug interaction: nodes represent drug; edges represent drug-drug interactions. Based on our assumption, we convert the prediction of DDI to link prediction problem, utilizing known drug node characteristics and DDI types to predict unknown DDI types. This work proposes a Graph Distance Neural Network (GDNN) to predict drug-drug interactions. Firstly, GDNN generates initial features for nodes via target point method, fully including the distance information in the graph. Secondly, GDNN adopts an improved message passing framework to better generate each drug node embedded expression, comprehensively considering the nodes and edges characteristics synchronously. Thirdly, GDNN aggregates the embedded expressions, undergoing MLP processing to generate the final predicted drug interaction type. GDNN achieved Test Hits@20=$0.9037\pm0.0193$ on the ogb-ddi dataset, proving GDNN can predict DDI efficiently.

Our code is available at https://github.com/zhf3564859793/GDNN.

## 1. Introduction

Combination medication is a common clinical treatment plan, especially for patients with muti-diseases. However, combination medication rises the risk of drug-drug interaction (DDI) contemporaneously. DDI refers to the pharmacological interaction between drugs, which lead to the enhancement or weakening of the drug efficacy. Furthermore, adverse drug reactions would be occurred. DDI causes massive dead patients as well as approximately 177 billion losses annually [1]. As a result, accurately predicting DDI becomes a clinically significant and urgent task to reduce both the risk and cost. Currently, conventional experiments in vivo and vitro can be used to identify DDIs, however, their shortages, laboratory limitations and unable afford high costs, can not be neglected. Therefore, computational prediction method is particularly central for DDI prediction.

*corresponding author

DDI prediction can be converted to the problem of node-node prediction in the graph. Expressly, we utilize complex relationships between drugs to form a graph, abstract drugs as the nodes; abstract drug-drug interaction as the edge connecting two nodes. Recently, deep learning is widely applied to solve problems of link prediction in graphs. Marinka et al. build a GCN architecture (Decagon) for predicting DDI [2]. Feng et al. combined GCN and DNN models to extract drug structural features from DDI networks to predict DDI[3]. In this work, based on graph neural networks framework, we utilize graph node-node distance information as node initial feature, comprehensively considering the nodes and edges characteristics synchronously to improve the accuracy of DDI prediction.

## 2. Problem Description

For a given DDI figure $G = \{V, E\}$, $V$ represents the collection of node in graph, including total $N$ nodes. Node feature matrix is represented by $X \in R^{N \times D}$. $E$ represents the collection of edges in graph. For a given drug, $x_i \in R^D, x_j \in R^D$, drug-drug interaction of $x_i$ and $x_j$ can be represented by $e_{i,j}$, that is $e_{ij} \in \{0, 1\}$. In above formula, $D$ represents the dimension of the node feature while $e_{ij}$ is referred to edge feature $Y \in \{0, 1\}^E$.

Since we convert DDI prediction to a link prediction problem, we assume there are some missing edges $Y^U$ in the graph, that is $Y = \{Y^U, Y^L\}$. With the premise of known node feature X and part known DDI link $Y^L$, our target is to predict unknown DDI type $Y^U$.

## 3. GDNN Framework

We propose the GDNN framework to solve DDI prediction problem. GDNN framework is consisted of three parts (Figure 1). The framework first generates node initial features through target method, then generate a node embedded representation via GDNN link predictor (Decoder part) to form relationship between two nodes. GDNN framework will be introduced in detail in the following parts.

## 3.1 Distance Encoding

As there is no node feature in graph ogb-ddi, we need to generate initialized features for the nodes. A current common method is to randomly generate node features through word embedding, however, this method neglects the structural information in the graph. To fully utilize graph information, Boling Li et al proposed to view node-node distance as node initial features, which sharply raise the accuracy of DDI prediction[4]. The disadvantage, compute node-node distances in graph, is still a tedious task. To simply calculation process, this work accepts target-based distance calculation method. We choose $k$ nodes as our target, then calculate the distance from every node to the chosen $k$ nodes. We use $X \in R^{N \times k}$ to represent generated distance feature matrix and distance feature matrix $X$ as initial node features in the graph.

Different selection of targets will lead to different initial node characteristics. As a result, we adopt the following three target selection methods. 1) Randomly select irredundant $k$ nodes from the graph as our targets, 2) Select the $k$ nodes with smallest node degree from the graph, 3) Select the $k$ nodes with the largest node degree from the graph

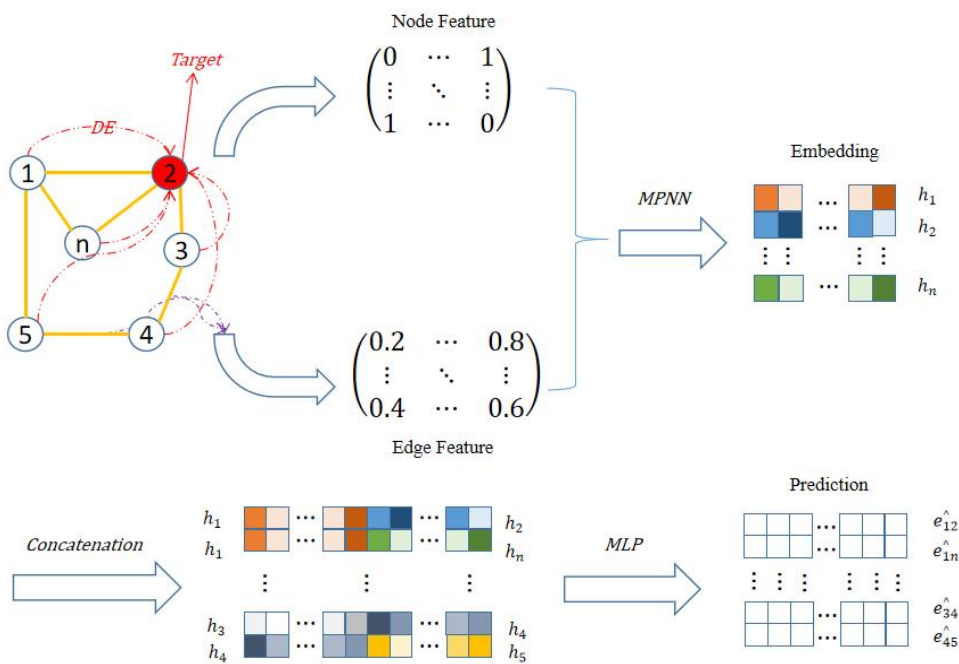

Figure 1. GDNN Framework

## 3.2 Messaging Passing Framework

Graph Neural Networks (GNNs) have already been used for connection prediction. To predict unknown drug-drug interaction in graph, GNN first learns all drug nodes embedding representation, then continue the following prediction according to these embedding representations. A classic node-based messaging frame work can be expressed as formula (1). Assume the hidden status of node $x_i$ in $t$ turn iterate is $h_i^t$. Its state at step $t+1$ is updated according to formula (2).

$$m_i^t = A(\{h_j^t\}_{j \in N(i)}) \tag{1}$$

$$h_i^{t+1} = U(h_i^t, m_i^t) \tag{2}$$

Among, $m_i^t$ is the message accepted by node in $t$ turn iterate, $N(i)$ represents neighbor node set of graph node $x_i$. $A(.)$ is message aggregation function. $U(.)$ is node update function.

In DDI graph, edge feature also shows important role, therefore, the messaging framework based on nodes neglect edge information, which will lead to lower prediction efficiency. We conclude edge feature in our work to solve the problem above. Thence, we use the following messaging framework:

$$m_i^t = A(\{f(e_{ij}).h_j^t\}_{j \in N(i)}) \tag{3}$$

$$h_i^{t+1} = U(h_i^t, m_i^t) \tag{4}$$

First, we find every neighbour edge of every node $x_i$, then we find another node connected by the neighbour edge. Secondly, we aggregate the feature vectors of the neighbor nodes and edges. Thirdly, we transfer the aggregated information $m_i^t$ to node $x_i$. Lastly, we use the update function to get the next iteration state $h_i^{t+1}$. The node status update rules are as follows:

$$h_i^{t+1} = W_1 h_i^t + W_2 \sum_{j \in N(i)} f(e_{ij}).h_j^t \tag{5}$$

In above formula, $f(.)$ is a MLP. The target of MLP is to transfer edge feature to an vector which can directly multiplied by the node eigenvector. The node embedding representation we get contains both the node and edge features in the graph.

The graph edge numbers, on the other hand, are far away enormous than the graph node numbers. As a result, the introduction of edge features will significantly increase the complexity of the algorithm. It also puts forward high requirements on the device video memory in the experiment. To solve this issue, we make a one-step modification to the above messaging framework[5]. Since the GraphSage framework can selectively continue neighbor sampling for target nodes, we sampling the edges between the two target nodes at the same time based on GraphSage framework basics. Then, we perform aggregation operations. Node feature update rules are as follow:

$$h_i^{t+1} = W_1 h_i^t + W_2.mean(\sum_{j \in N(i)} W_3 e_{ij} + h_j^t) \tag{6}$$

Carrying out message passing like formula (6) can greatly reduce the number of collected edges, thereby significantly reducing the algorithm complexity.

### 3.3 Link Predictor

After obtaining the embedding representation of the node, we adopt MLP as the link predictor, the specific operation is as follows:

$$\hat{e_{i.j}} = MLP\big(h_i \times h_j\big) \tag{7}$$

For an edge, we multiply its two endpoints embedding expressions as the edge feature vector. Then, we process the MLP to obtain the predicted value of the edge type $\hat{e_{i.j}}$.

## 4. Experiment

### 4.1 Dataset

This article utilizes the ddi dataset officially provided by ogb[6]. The Ogb-ddi dataset is a homogeneous, weightless, and undirected graph, where each node represents a drug, and each edge represents a drug-drug interaction.

### 4.2 Experiment Setting and Evaluation Indicators

We compare GDNN model effect with other baseline models like GNN, Graphsage, JKNet. We set up ablation experiments to build a GDNN⁻ model (without adding edge features) to verify the effectiveness of edge features

We use the Hits@20 indicator provided by the official ogb.

### 4.3 Hyperparameters Setting

We study the four hyperparameters effects on our final prediction results, encoder, decoder hidden layer dimensions, node feature matrix dimensions, as well as target selection methods.

#### 4.3.1   The Best Node Dimension

The the node feature matrix dimension can have a enormous impact on model performance. Since GDNN requires quite long computing time, we conducted corresponding comparative experiments on the GDN N⁻ model. The specific results are shown in Figure 2.

From Figure 2, the DDI model prediction accuracy increases first with the increase of node dimension. It reaches the maximum value at 512 dimensions, then, increasing the node dimension will degrade the performance of the model. As a result, we choose 512 as our optimal node dimension.

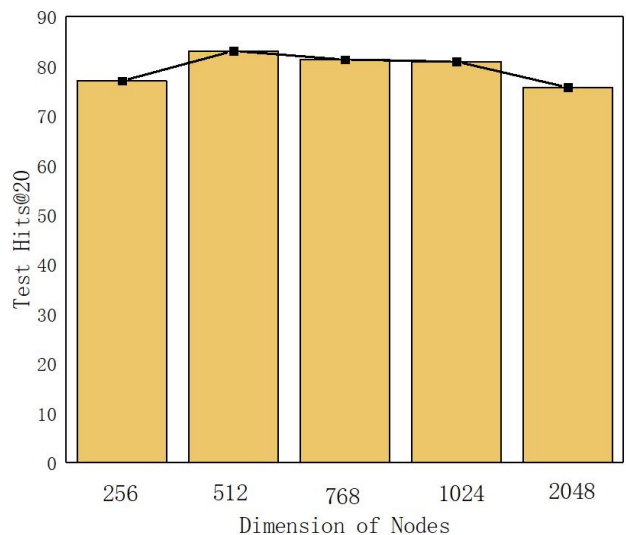

Figure 2. Influence of node feature dimension on model performance

### 4.3.2 The way to Choose Target

We investigated the impact of three different target selection approaches on model outcomes. We conducted corresponding comparative experiments on the GDNN⁻ model. The specific results are shown in Figure 3.

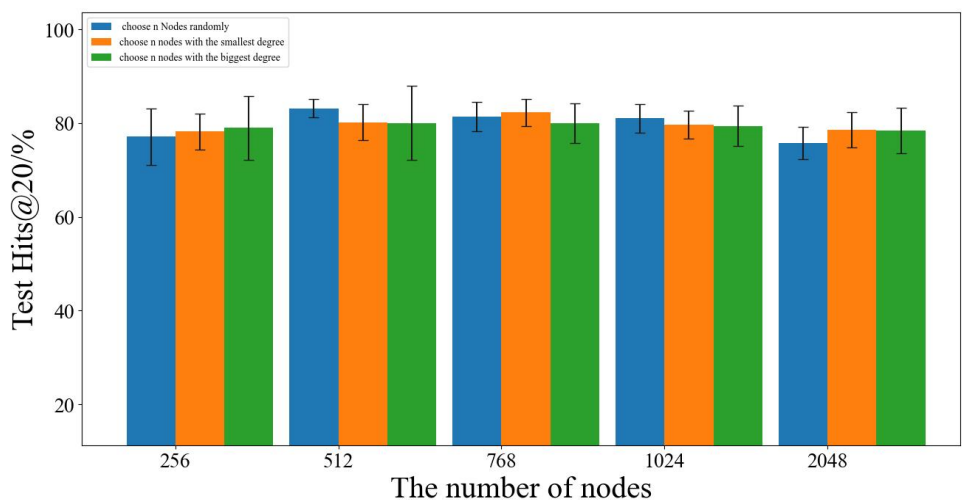

Figure 3.The Affection of Different Targets Selection Methods

In most cases, randomly selecting nodes and selecting nodes with the smallest node degree as targets can achieve better experimental results. This tendency maybe complicated with the node feature matrices generated by these two methods. Our node feature dimensions are all 512 in GDNN, therefore, we randomly select 512 nodes as targets to generate the initial node features.

### 4.4 Results

The specific experimental results are shown in Table 1. It is evident that our GDNN model has achieved the best results.

Table 1. The Results on Ogb_ddi Dataset

| Method | Validation Hits@20 | Test Hits@20 |
|---|---|---|
| GNN | $0.5550 \pm 0.0208$ | $0.3707 \pm 0.0507$ |
| Graphsage | $0.6262 \pm 0.0037$ | $0.5390 \pm 0.0474$ |
| JKNet | $0.6776 \pm 0.0095$ | $0.6056 \pm 0.0869$ |
| DEA + JKNet | $0.6713 \pm 0.0071$ | $0.7672 \pm 0.0265$ |
| GraphSAGE+anchor distance | $0.8239 \pm 0.0437$ | $0.8206 \pm 0.0298$ |
| **GDNN$^-$** | **$0.7528 \pm 0.0181$** | **$0.8316 \pm 0.0192$** |
| CFLP (w/ JKNet) | $0.8405 \pm 0.0284$ | $0.8608 \pm 0.0198$ |
| GraphSAGE + Edge Attr | $0.8044 \pm 0.0404$ | $0.8781 \pm 0.0474$ |
| **GDNN** | **$0.8599 \pm 0.0286$** | **$0.9037 \pm 0.0193$** |

## 5. Conclusion

In conclusion, GDNN uses the target method to generate initial features for the nodes in the graph. We improve the framework of GraphSage, comprehensively consider the features of nodes and edges in the graph. We better generate each drug node embedding, and reduce the algorithm complexity. The experimental results prove that the GDNN model has achieved good results in DDI prediction.